\documentclass[preprint,12pt]{elsarticle}

\usepackage{amsmath, hyperref, url}
\usepackage{multirow}
\usepackage{doi}
\usepackage{nomencl}
\usepackage{soul}
\soulregister{\citep}7
\makenomenclature

\journal{Computers and Electronics in Agriculture}

\begin{document}
\begin{frontmatter}

\title{MOT-DETR: 3D Single Shot Detection and Tracking with Transformers to build 3D representations for Agro-Food Robots}

\author[inst1]{David Rapado-Rincon}

\affiliation[inst1]{organization={Agricultural Biosystems Engineering, Wageningen University \& Research}, 
            addressline={P.O. Box 16}, 
            city={Wageningen},
            postcode={6700 AA}, 
            country={The Netherlands}}
\author[inst1]{Henk Nap}
\author[inst2]{Katarina Smolenova}
\affiliation[inst2]{organization={Business Unit Greenhouse Horticulture, Wageningen University \& Research}}
\author[inst1]{Eldert J. van Henten}
\author[inst1]{Gert Kootstra}

\begin{abstract}
In the current demand for automation in the agro-food industry, accurately detecting and localizing relevant objects in 3D is essential for successful robotic operations. However, this is a challenge due the presence of occlusions. Multi-view perception approaches allow robots to overcome occlusions, but a tracking component is needed to associate the objects detected by the robot over multiple viewpoints. Most multi-object tracking (MOT) algorithms are designed for high frame rate sequences and struggle with the occlusions generated by robots' motions and 3D environments. In this paper, we introduce MOT-DETR, a novel approach to detect and track objects in 3D over time using a combination of convolutional networks and transformers. Our method processes 2D and 3D data, and employs a transformer architecture to perform data fusion. We show that MOT-DETR outperforms state-of-the-art multi-object tracking methods. Furthermore, we prove that MOT-DETR can leverage 3D data to deal with long-term occlusions and large frame-to-frame distances better than state-of-the-art methods. Finally, we show how our method is resilient to camera pose noise that can affect the accuracy of point clouds. The implementation of MOT-DETR can be found here: https://github.com/drapado/mot-detr
\end{abstract}


\end{frontmatter}
\printnomenclature

\section{Introduction}
The agro-food industry is facing increasing challenges due to a growing and more affluent population, alongside a decreasing labor force. Automation, particularly robotics, is considered a key solution but encounters issues in these environments, primarily in terms of robotic perception and interaction because of factors like high occlusion and variation \citep{kootstra_selective_2021}.

\begin{figure*}[ht]
    \centering
    \includegraphics[width=\textwidth]{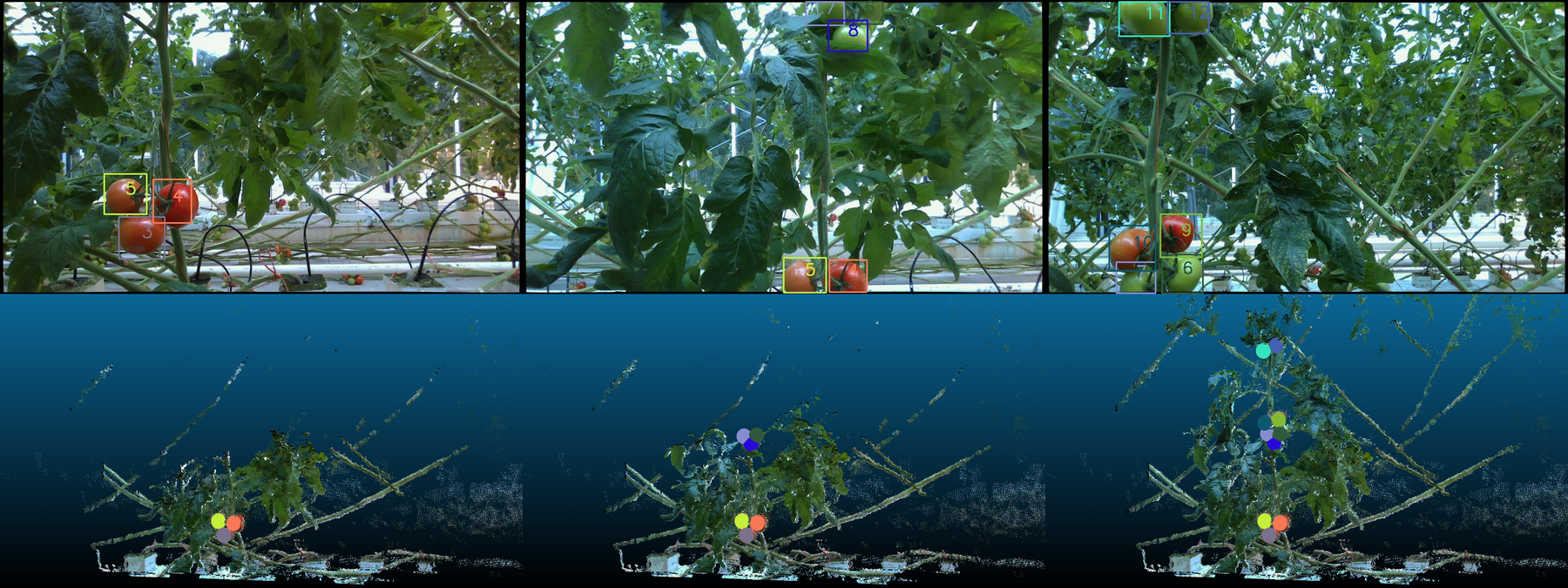}
    \caption{Example of the tracking and 3D representation results over five frames of a sequence. The top row shows the viewpoint color image and the bounding boxes of the tracked tomatoes, with their corresponding tracking ID. The bottom row contains the merged point cloud from the first frame. Additionally, overlapping the point cloud there are spheres that represent each tracked tomato in 3D.}
    \label{fig:qualitative}
\end{figure*}

An accurate and efficient representation of the robot's operating environment, including objects' locations and properties, is essential for successful robot operation in these environments \citep{crowley_dynamic_1985,elfring_semantic_2013}. Traditional robotic systems using single-view data, like an image, fail to capture all relevant details due to the high occlusion levels of agro-food environments. This leads to errors in detecting and locating objects \citep{arad_development_2020}. Representations that incorporate multiple views have the potential to improve detection and localization even in highly occluded conditions. Here, active perception methods can play an important role in selecting the most optimal viewpoints \citep{burusa_efficient_2023}. However, developing representations from multi-view perception necessitates robust data association. Data association involves linking detections from different viewpoints to the correct object representations, ensuring that each detected object is consistently tracked and represented across all views. In the context of an object-centric representation, this task is known as multi-object tracking (MOT\nomenclature{\(MOT\)}{Multi-Object Tracking}). MOT is crucial as it helps in correlating information about the same object observed from various perspectives over time, despite occlusions and changes in the object's appearance or position \citep{elfring_semantic_2013, wong_data_2015, persson_semantic_2020, rapado-rincon_development_2023}.

Performing 3D MOT in complex agro-food environments remains a challenge due to high levels of occlusion and sensor noise. These factors complicate the task of consistently and accurately associating detections with their respective objects. Yet, the accuracy of the MOT algorithm directly impacts the quality of the overall representation, which is vital for the operational success of robotic systems in such environments \citep{rapado-rincon_development_2023}.

In this paper, we focus on addressing the problem of 3D data association and 3D MOT that arises in multi-view perception. We propose a method to build representations that can be used by robots through a novel approach to perform object detection and MOT in 3D.
Our method enables the use of multi-view perception to deal with occlusions by detecting and associating the relevant objects in the present viewpoint with the already detected objects from previous viewpoints. This allows robotic systems to build an accurate representation in occluded environments. Figure \ref{fig:qualitative} shows an example of the resulting tracking and 3D representation. 

\section{Related work and contributions}
In MOT, different strategies are used. \textit{Two-stage methods}, like Simple Online and Real-time Tracking (SORT) \citep{bewley_simple_2016} and DeepSORT \citep{wojke_simple_2017}, have been popular in the last years. Two-stage methods typically use a deep-learning object detector, and an association algorithm that associates upcoming detections with previous ones. SORT passes the 2D position of the detected objects to an association method based on a Kalman filter and a Hungarian algorithm. DeepSORT extended SORT by adding a feature extraction network to generate re-identification (re-ID) features that help the association algorithm. \textit{Single-stage methods} combine detection and re-ID in one step, making them more efficient. However, the association is still performed separately. A recent example of single-stage is FairMOT \citep{zhang_fairmot_2021}. In \textit{recurrent methods}, the detection of objects and its association over multiple frames is performed through the same network. Recent examples of recurrent methods are Trackformer \citep{meinhardt_trackformer_2022} and MOTR \citep{zeng_motr_2022}. In both methods, the tracking is performed end-to-end by a single-stage network that leverages the transformer architecture to detect objects of DETR \citep{carion_end--end_2020} and to track them over consecutive frames. The transformer architecture of DETR allows it to reason about the objects and the relations between themselves and the image context. This helps the helps the model in preventing double detections as proven by \citep{carion_end--end_2020} and can be very useful for MOT applications. Recurrent methods based on transformers eliminate the need for complex handcrafted association components, providing a more streamlined approach to MOT. However, their recurrent-like approach complicates the training process compared to algorithms single-stage algorithms \citep{zeng_motr_2022}.

In agro-food environments, MOT has been employed for tasks like crop monitoring and fruit counting \citep{halstead_fruit_2018, kirk_robust_2021, halstead_crop_2021, villacres_apple_2023}. However, occlusions can reduce a tracking algorithm's performance \citep{rapado-rincon_development_2023}. This indicates the need for more powerful tracking algorithms. Even though there exist MOT algorithms with more novel and potentially better tracking performance as indicated earlier, most agro-food tracking approaches are still based on SORT or DeepSORT as they require less data and are less complex to train.

Above-mentioned MOT algorithms, like DeepSORT \citep{wojke_simple_2017}, FairMOT \citep{zhang_fairmot_2021} and MOTR \citep{zeng_motr_2022}, were designed to work in video sequences with high frame rates, where the difference between frames is small. This tends to simplify tracking. However, in robotic applications, differences between frames are not always small due to the 3D nature of robot' motions and environments. This results in sequences with more prevalent and complex occlusions, and drastic perspective changes. Using 3D data for MOT can solve this issue by providing additional information to distinguish objects more effectively \citep{rapado-rincon_development_2023, rapado-rincon_minksort_2023}. Nevertheless, to overcome some of the challenges of 3D environments and motions, algorithms that can better utilize 3D data for MOT are needed. 

In this paper we present a novel method, MOT-DETR (Multi-Object Tracking and DEtection with TRansformers), to perform MOT tracking in complex agro-food environments. Our contributions are as follows:
\begin{itemize}
\item A novel deep learning method, MOT-DETR, which is an adaptation of the transformer-based architecture developed by DETR \citep{carion_end--end_2020} with an extra re-ID output to perform MOT in a single-stage manner. MOT-DETR makes use of the capabilities of the transformer architecture to perform MOT without the complexity of recurrent-based methods. It improves against a single-shot detection and tracking state-of-the-art method, FairMOT \citep{zhang_fairmot_2021}, independently of the frame-to-frame distance. 
\item A method of using 3D information by leveraging the capabilities of self- and cross-attention of the transformer architecture to merge information from color images and point clouds to improve the MOT performance of MOT-DETR.
\item A novel method to generate random synthetic tomato plant 3D models using using an L-system-based formalism that can be used to render viewpoints and generate training and evaluation sequences. The combination of a small real set with a large synthetic set generated using this method during training greatly improves the performance of MOT-DETR.
\end{itemize}

\section{Materials and Methods}
\begin{figure*}[ht]
    \centering
    \includegraphics[width=\textwidth]{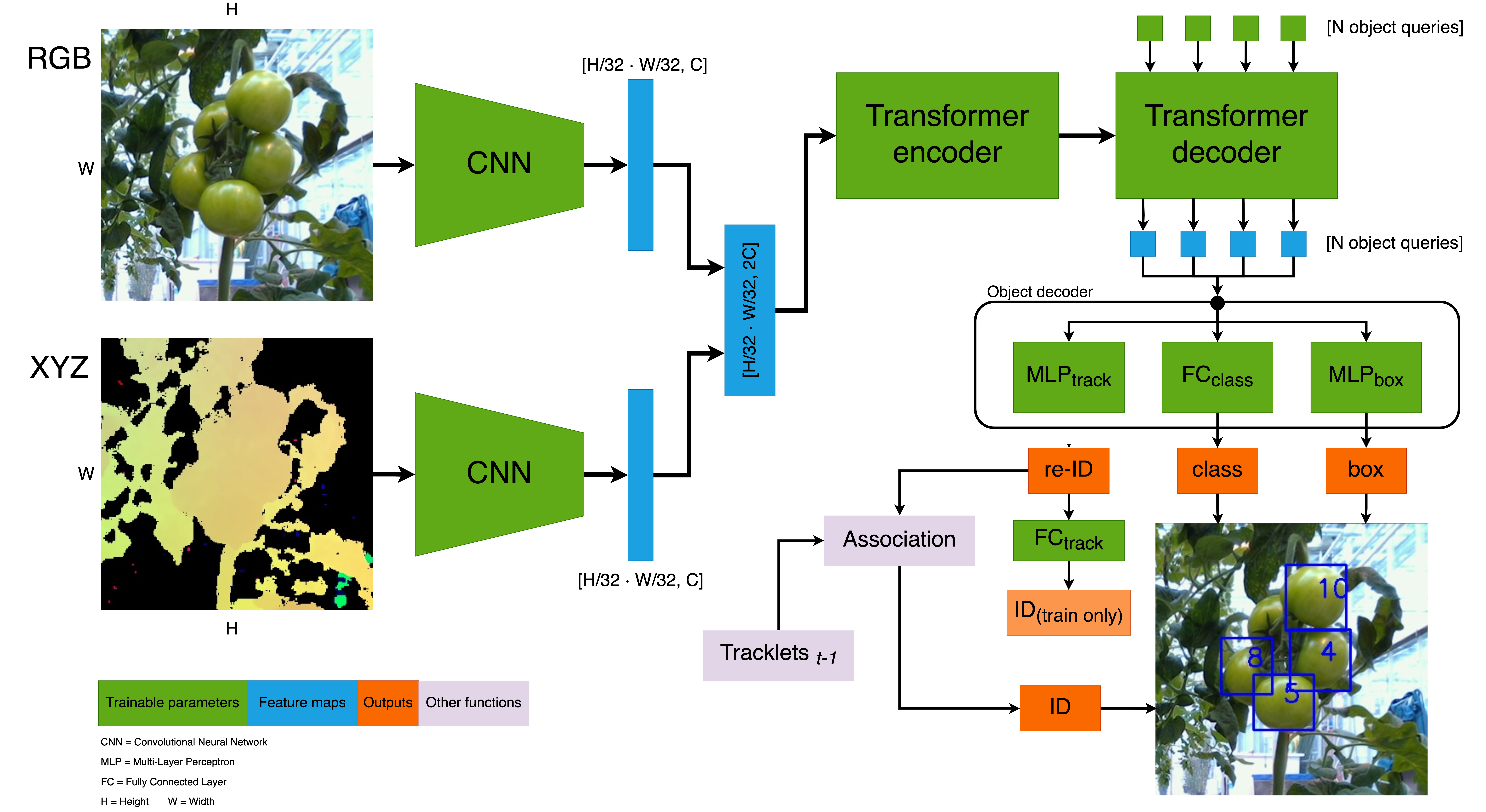}
    \caption{MOT-DETR architecture: color images and point clouds are processed using independent CNNs, then the resulting feature maps are concatenated and passed to a transformer. The transformer predicts a set of bounding boxes with corresponding class and re-ID features. The re-ID features are passed to a data association algorithm that generates the ID of every detection.}
    \label{fig:alg}
\end{figure*}

In this work, we present MOT-DETR, a MOT algorithm that uses the single-shot detection and tracking approach of FairMOT \citep{zhang_fairmot_2021} but with the detection architecture of DETR \citep{carion_end--end_2020}. We further enhanced MOT-DETR to process simultaneously 2D images and 3D point clouds. For every viewpoint that the robot collects, MOT-DETR predicts the following outputs per detected object: a 2D bounding box, the object class, and re-ID features. The re-ID features are then used in a data association process that associates the objects detected at the current viewpoint with previously detected objects, also referred to as tracklets. The data association is done by using the re-ID features of newly detected and previous objects to build a cost matrix using the cosine distance. This cost matrix is then passed to a Hungarian algorithm that generates the associations for tracking. The architecture of MOT-DETR can be seen in Figure \ref{fig:alg}.

\subsection{Data pre-processing}
For each viewpoint the robot collects, MOT-DETR takes as input a color image and its corresponding structured 3D point cloud. The point cloud is transformed using the known transformation between the camera and the robot's world coordinate system. In this way, all point clouds collected over time have the same coordinate system. Next, each point cloud is transformed into a normalized image by workspace-defined limits for each Cartesian axis. These limits are set based on the expected volume of both real and synthetic tomato plants, as well as the working space limits of the robot arm. This step also removes points in space that are outside of the target area of the robot. 

\subsection{MOT-DETR architecture}
Both color image and point cloud are processed through two independent convolutional neural networks (CNNs\nomenclature{\(CNN\)}{Convolutional Neural Network}). We used two independent ResNet34 CNNs for our implementation. The CNNs were pre-trained on ImageNet. The output of the CNNs is then flattened, and a fully connected layer is used to reduce the dimension of the feature to $C$ channels. After this, both flattened maps are concatenated, increasing the feature dimensions to $2C$. Following the architecture of DETR \citep{carion_end--end_2020}, the concatenated feature map is passed to a transformer encoder, which applies attention mechanisms to focus on different parts of the image.
The output of the transformer encoder is then passed to the transformer decoder. For this work, both transformer encoder and decoder have three layers. As in DETR, the transformer decoder decodes a set of $N$ object queries using self- and encoder-decoder attention mechanisms. These object queries are independent learnable vectors that are used to query the transformer decoder for the presence of objects in the image. The $N$ object queries are transformed into $N$ output embeddings, which are then passed through a set of prediction heads. There are three heads: one for class prediction, one for bounding box prediction and one for re-ID features. The class prediction head outputs the probabilities of each class for each object query. However, not all $N$ embeddings represent a true object. Most of them correspond to background. To differentiate them, the class "background" is used together with the different object classes present in the dataset. The classification head then predicts, for each output embedding, if it is an object of any of the target classes or if it is not an object at all. The bounding box prediction head outputs the coordinates of the 2D bounding box for each object query, and the tracking head predicts the unique ID of each object at training time and a set of re-ID features at inference time.

Apart from the default version of MOT-DETR, which takes 2D images and 3D point clouds as input (3D version), we also implemented a version that takes only 2D images as input (2D version). This version has only one CNN, whose resulting feature map is passed to the transformer network. We doubled the output size of the fully connected layer after the CNN to keep the size of the transformer network the same between the 3D and the 2D versions. Consequently, the output size of this layer is $2C$ for the 2D version instead of $C$ as in the 3D version.

\subsection{Training}
MOT-DETR performs three tasks, defined by the three prediction heads: object classification, bounding box prediction, and object tracking. Each task has its own loss function that contributes to the training process.

A crucial step to train an object detection network is associating each ground truth object with one of the network predictions. We use the same approach as \citep{carion_end--end_2020}, the Hungarian matching. The cost matrix for the Hungarian algorithm is computed using the object detection loss. As in DETR \citep{carion_end--end_2020}, the object detection loss corresponds to the sum of the cross entropy (CE\nomenclature{\(CE\)}{Cross Entropy}) loss, L1 loss and the Generalized Intersection over Union (GIoU\nomenclature{\(GIoU\)}{Generalized Intersection over Union}) between predicted objects and ground truth objects. This approach ensures that each ground truth object is matched with one prediction, and this matching minimizes the total object detection loss. 

Once predictions have been associated to ground truth objects, the total loss can be calculated as
\begin{equation}
L_{\text{total}} = \frac{1}{2} (e^{w_1} L_{\text{det}} + e^{w_2} L_{\text{id}} + w_1 + w_2)
\end{equation}
where $L_{\text{det}}$ corresponds to the object detection loss as defined by DETR \citep{carion_end--end_2020}, $L_{\text{id}}$ is the re-ID loss as defined in FairMOT \citep{zhang_fairmot_2021}, and $w_1$ and $w_2$ are learnable parameters that balance the two tasks of object detection and re-identification. 

We trained MOT-DETR using the AdamW optimizer \citep{loshchilov_decoupled_2019} with a learning rate of $1 \times 10^{-4}$ during 40 epochs. Using an RTX 4090, this resulted in a training time of approximately 15 hours.

\subsection{Inference and tracking}
At inference time, we discard the tracking ID predicted by the network, and we use the re-ID features. These features, are fed into a data association algorithm which uses the Hungarian algorithm to associate objects over multiple viewpoints. At every viewpoint, the cost matrix between the existing tracklet features, and the newly detected object features is calculated using the cosine distance. Then the Hungarian algorithm is used to select the best assignments between existing objects and new detections. Detections that were not associated with any existing object initiate a new object tracklet. Tracklets that have not been associated with any detection remain.

The 3D location of every tomato is tracked and updated with a Kalman filter as in our previous work \citep{rapado-rincon_development_2023}. When a tomato is detected, its position updates the Kalman filter state. This position is derived by filtering the point cloud within the tomato's bounding box and averaging the points.

\begin{table}[htbp]
\centering
\caption{Distribution of plants and images in train, validation and test splits.}
\resizebox{\linewidth}{!}{
    \begin{tabular}{cccc}
    \hline
    \textbf{Type} & \textbf{Split} & \textbf{\# Plants} & \textbf{\# Viewpoints} \\ \hline
    \multirow{2}{*}{Real} & Train / Validation & 4 & 3,570 / 630 \\
    & Test & 1 & 1,200 \\ \hline
    \multirow{2}{*}{Synthetic} & Train / Validation & 45 & 47,500 / 2,500 \\
    & Test & 5 & 5,000 \\ \hline
    \end{tabular}
}
\label{tab:dataset}
\end{table}

\begin{figure}[ht]
    \centering
    \includegraphics[width=\textwidth]{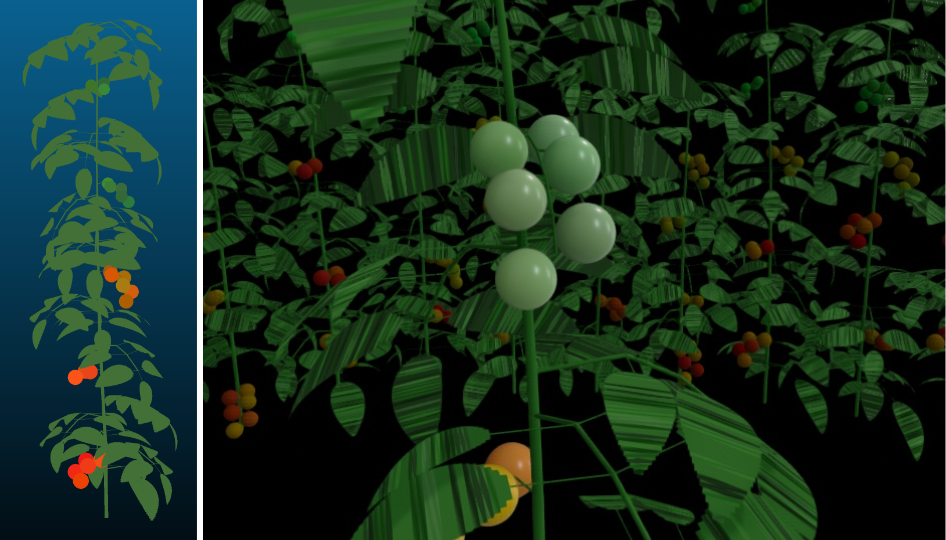}
    \caption{\textbf{Left.} Example of a synthetic plant model. \textbf{Right.} Example of a generated viewpoint. Extra plants are added in the background to increase the complexity of the scene.}
    \label{fig:syn_plant}
\end{figure}

\subsection{Data}
Deep neural networks require large amounts of data to train. Applied fields, such as agro-food, have worked around this problem by using pre-trained models. However, these practices are less common in 3D applications, where widely standardized methods like CNNs and large general datasets are less available. To solve this problem, we developed a method to generate random synthetic tomato plant 3D models. 

The 3D plant models were generated using an L-system-based formalism in the modeling platform GroIMP, v1.6 \citep{hemmerling_rule-based_2008}. The formalism allows to specify individual organs of the plant, like fruit or leaf, their geometry, and connections between them. The L-system was parameterized by reading in actual morphological data of a complete plant. Morphological data was obtained from measurements in young greenhouse-grown tomatoes. The architecture of these plants is relatively simple, made by a repetitive pattern of three leaves and one truss with fruits. To generate multiple plants, randomness was added to individual organ traits: leaf angle, leaf length and width, internode length, number of tomatoes per truss, and size of the tomatoes in each truss. In addition, a substantial set of real images was collected.

An example of a synthetic plant model is shown in Figure \ref{fig:syn_plant}-left. From the 3D models, we rendered color images and point clouds from random viewpoints using Open3D \citep{zhou_open3d_2018}. For each viewpoint, the position of the camera was sampled randomly inside of a cylinder whose center was approximately the plant stem. The camera would aim to a random point approximately around the stem of the plant. This process would result in viewpoints with five degrees-of-freedom (DoF\nomenclature{\(DoF\)}{Degrees of Freedom}). An example of the rendered RGB\nomenclature{\(RGB\)}{Reg Green Blue} image from a viewpoint can be seen in Figure \ref{fig:syn_plant}-right. In total, we generated 50 different plant models, and rendered 1,000 viewpoints for each of them, resulting in 50,000 viewpoints.

\begin{figure}[ht]
    \centering
    \includegraphics[width=\textwidth]{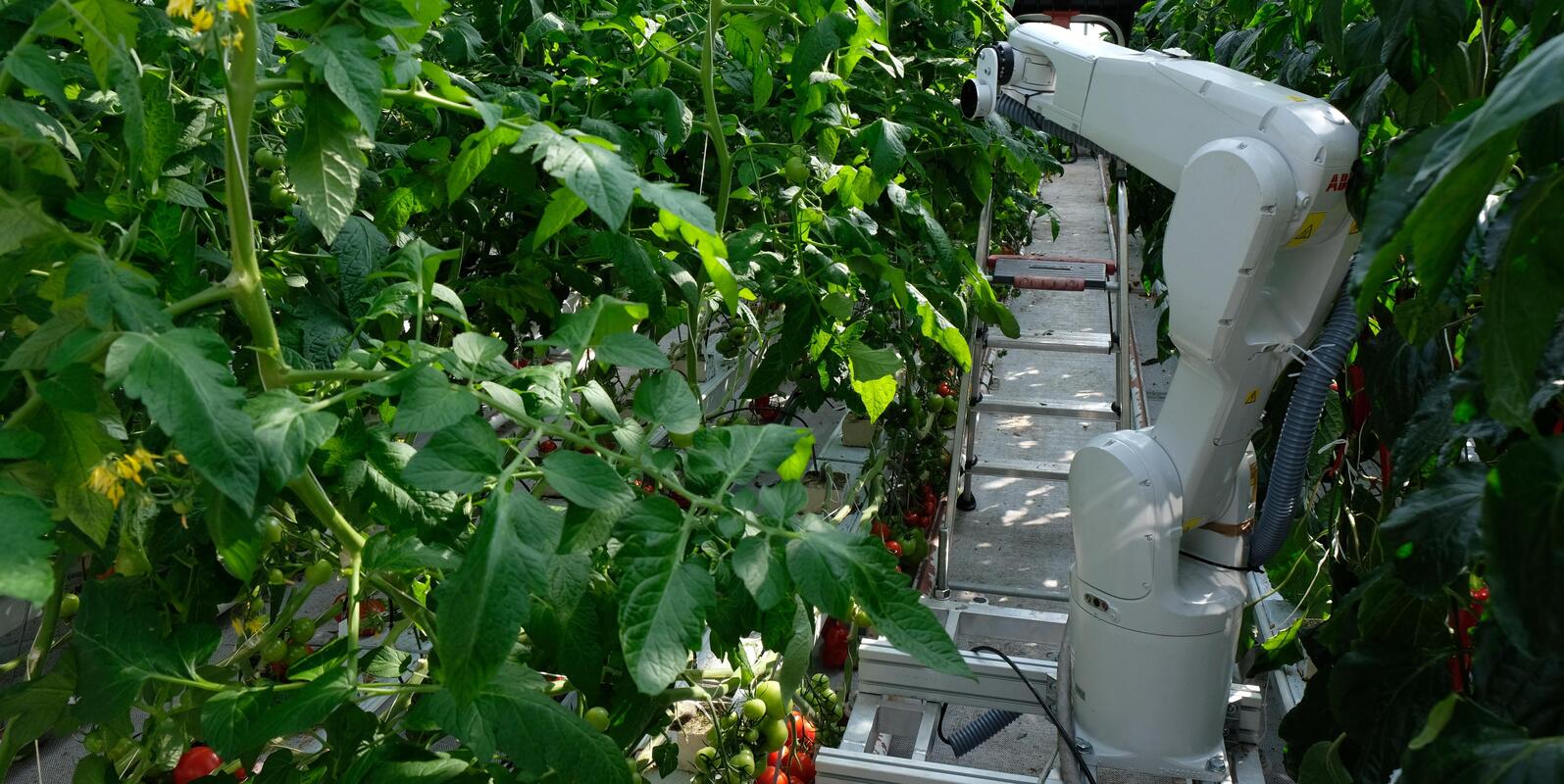}
    \caption{Robotic system used for data collection. The robot arm ABB IRB1200 is mounted over a mobile platform that allows motion over the greenhouse row rails. A Realsense L515 camera is mounted on the end-effector.}
    \label{fig:robot_greenhouse}
\end{figure}

\begin{figure}[ht]
    \centering
    \includegraphics[width=\textwidth]{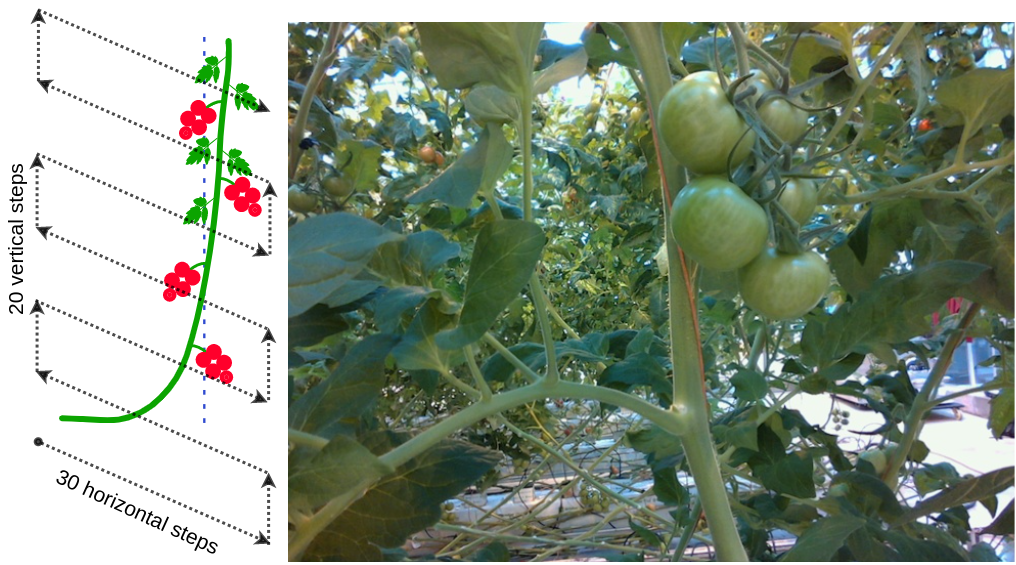}
    \caption{\textbf{Left.} Illustration of the path followed by the robot to collect viewpoints of real plants. \textbf{Right.} Example of a viewpoint in a real plant.}
    \label{fig:real_plant}
\end{figure}

Furthermore, a dataset using five real plants from a tomato greenhouse was collected using the system shown in Figure \ref{fig:robot_greenhouse}. The arm is an ABB IRB1200, and it is mounted on a rolling platform that can move over the heating pipces of the greenhouse. An Intel Realsense L515 RGB-D camera is mounted in the end effector of the robot. Per plant, viewpoints were collected using a two-DoF planar motion sequence in front of each plant at distances 40 cm, 60 cm, or both. In total, 5,400 viewpoints from real plants were collected. Figure \ref{fig:real_plant} shows the diagram of the data collection path in real plants on the left, and an example of an image viewpoint on the right. The resolution of the images is \textit{960x540} pixels.

To train and evaluate MOT-DETR, we divided both synthetic and real datasets into train, validation and test splits as shown in Table \ref{tab:dataset}. To prevent plants from being seen by the network at train time and during the experiments, train and validation splits came from the same pool of plants, while test splits were generated from different plants. We trained MOT-DETR with the combined real and synthetic train sets.

At training time, data augmentation was used for both the images and the point clouds. Color images were augmented using brightness, contrast, saturation and hue changes. Point clouds were augmented by adding six DoF noise to the camera pose transformation. A Gaussian distribution with a mean of zero and a standard deviation of 0.005 was used to augment the camera pose. Furthermore, both color images and point clouds were augmented by performing random cropping.

\section{Results and Discussion}
In this section we present the evaluation of the performance of MOT-DETR in different scenarios. We evaluate the detection performance and inference speed of MOT-DETR. We also compare our default MOT-DETR with 2D and 3D inputs (3D) against a version with only 2D input (2D), and a state-of-the-art MOT algorithm: FairMOT \citep{zhang_fairmot_2021} on different type of sequences with real and synthetic data. Furthermore, since all point clouds are transformed into the robot coordinate system using the camera pose of each viewpoint, we will show the effect of noise in the camera pose transformation in the performance of MOT-DETR.

\begin{table}[htbp]
\centering
\caption{Number of parameters, inference speed, and detection accuracy of MOT-DETR with 3 and 6 input channels.}
\resizebox{\linewidth}{!}{
    \begin{tabular}{cccccccc}
    \hline
    \textbf{Model} & \textbf{\# CNN} & \textbf{\# Transformer} & \textbf{\# Total} & \textbf{FPS} & \textbf{mAP}$_{real}$ & \textbf{mAP}$_{syn}$ \\ \hline
    MOT-DETR-3D & 42.6 M & 18 M & 60.6 M & 43.11 & 85.88 & 83.61 \\
    MOT-DETR-2D & 21.3 M & 18 M & 39.3 M & 61.21 & 82.20 & 85.65\\ \hline
    \end{tabular}
}
\label{tab:AP}
\end{table}

\subsection{Detection performance and Inference speed}
We studied the detection performance of the 2D and 3D versions of MOT-DETR. Table \ref{tab:AP} shows the number of parameters, inference speed, and the mean average precision (mAP\nomenclature{\(mAP\)}{Mean Average Precision}) of our two variants on both real and synthetic test sets. MOT-DETR-3D contains approximately 20 M more parameters than MOT-DETR-2D. Consequently, the inference time of MOT-DETR-3D is slower than its 2D version. Nevertheless, even MOT-DETR-3D is able to run at 43.11 frames per second on a Nvidia RTX 4090 GPU\nomenclature{\(GPU\)}{Graphical Processing Unit}. This is sufficient speed for most robotic applications.

The detection peformance, evaluated as mean average precision (mAP), of MOT-DETR-3D is three points larger on the real data, but two points lower on the synthetic data, compared to MOT-DETR-2D. This might be due to the greater uniformity of tomatoes in the synthetic data, facilitating their distinction from the background in the 2D data compared to the real dataset (see Figure \ref{fig:syn_plant}-right and Figure \ref{fig:real_plant}-right). Consequently, the additional insights offered by the 3D data do not significantly enhance the model's performance. In contrast, for the real data, leveraging 3D information can potentially augment the network's ability to better detect tomatoes.

Despite multi-tasking algorithms often exhibiting marginally inferior performance compared to single-task ones \citep{zhang_fairmot_2021}, our detection and re-identification network matches the performance of prevalent detection-only networks for tomatoes like the Mask R-CNN algorithm used by \citep{afonso_tomato_2020}, that achieved mean average precision values larger than 0.82. Even though they used a different dataset as the one collected for this work, their images have similar camera perspectives and tomato plants.

\subsection{Tracking performance}
The problem of object tracking in robotics is not always similar to object tracking in videos. Change of perspective and long-term occlusions are more common in robotic applications. To study the performance of our algorithm in multiple type of sequences, we defined three different experiments:

\begin{itemize}
    \item \textbf{Real-Sort.} For our real test set, we have two sequences of 600 frames at two distances from the plant. For each sequence, we selected 100 random viewpoints. Then we sorted the viewpoints with the objective of minimizing frame-to-frame distance. This generates a 100 frame sequence similar to a video sequence with low frame rate. 
    \item \textbf{Real-Random.} Similarly to the previous experiment, 100 viewpoints were selected out of the pool of viewpoints per plant. However, they are not sorted in this case. This generates a sequence where jumps between frames are larger, and objects are occluded during several frames.
    \item \textbf{Synthetic-Random.} Our synthetic dataset was recorded with more degrees of freedom than our real data. Consequently, we designed an experiment to evaluate the performance of our algorithm when 100 random and unordered frames are selected out of the pool of viewpoints of our synthetic test plants.
\end{itemize}

\begin{figure}[ht]
    \centering
    \includegraphics[width=\textwidth]{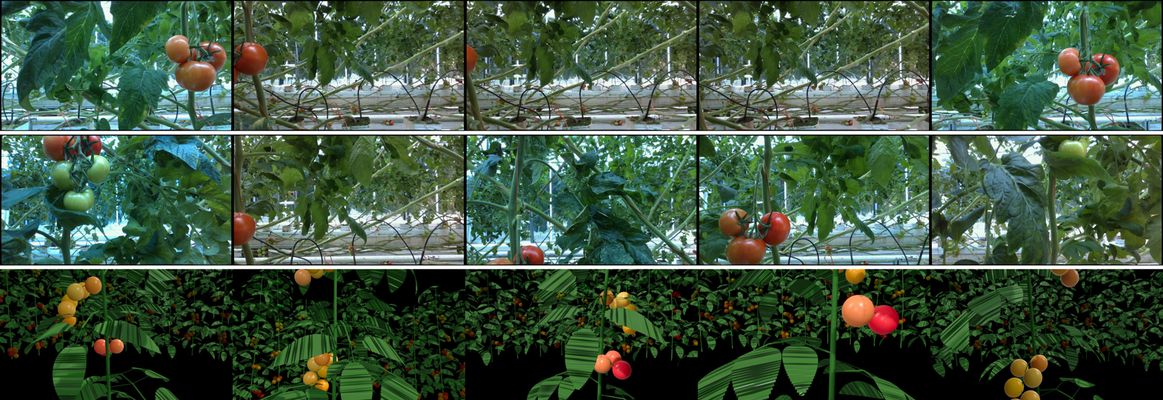}
    \caption{First five frames of one of the sequences of each experiment: Real-Sort (top row), Real-Random (middle row), and Synthetic-Random (bottom-row).}
    \label{fig:seqs}
\end{figure}

We evaluated the tracking accuracy using High Order Tracking Accuracy (HOTA\nomenclature{\(HOTA\)}{High Order Tracking Accuracy}), with its sub-metrics Localization Accuracy (LocA\nomenclature{\(LocA\)}{Localization Accuracy}), Detection Accuracy (DetA\nomenclature{\(DetA\)}{Detection Accuracy}) and Association Accuracy (AssA\nomenclature{\(AssA\)}{Association Accuracy}); and with Multi-Object Tracking Accuracy (MOTA\nomenclature{\(MOTA\)}{Multi-Object Tracking Accuracy}) and its sub-metric ID Switches (IDSW\nomenclature{\(IDSW\)}{ID Switches}). Each experiment was repeated five times by selecting a different set of 100 random viewpoints. A t-test was then used to assess the significance of the differences between the models. The first five frames from a sequence in each experiment are illustrated in Figure \ref{fig:seqs}.

We compared three models: MOT-DETR-2D, MOT-DETR-3D, and the state-of-the-art tracking algorithm FairMOT \citep{zhang_fairmot_2021}. FairMOT is a single-shot detection and tracking algorithm designed to track objects in video sequences, it has been shown to outperform commonly used tracking algorithms in agriculture like SORT \citep{bewley_simple_2016} and DeepSORT \citep{wojke_simple_2017}. The key architectural differences between our proposed method MOT-DETR and FairMOT are its architecture being based on CenterNet, which is a standard CNN, and the use of only 2D data. Furthermore, for this experiment, we used a pre-trained FairMOT network on large object detection and tracking datasets, and we fine-tuned it using our real tomato training set. Since our synthetic dataset was obtained using completely random and independent viewpoints with 5DoF, it could not be adapted in any way to the video sequences that FairMOT was designet to. Therefore, we did not trained nor evaluate FairMOT with our synthetic dataset.

In the Real-Sort experiment, Table \ref{tab:combined} reveals MOT-DETR-2D to be nearly on par with FairMOT. However, MOT-DETR-3D clearly outperforms any model using only 2D data as input. This is expected as the use of 3D data provides relevant information for tracking objects that look similar, like tomatoes. FairMOT \citep{zhang_fairmot_2021} only outperforms MOT-DETR in LocA, potentially due to its prior training on a vast object detection dataset, unlike MOT-DETR.

For the Real-Random experiment, the performance difference between MOT-DETR-3D and both of the 2D algorithms becomes larger. This is due to the performance of MOT-DETR-2D and FairMOT decreasing more than MOT-DETR-3D when the distance between viewpoints is larger, showing that MOT-DETR-3D can successfully use 3D data to improve tracking accuracy in challenging sequences. Furthermore, it can be seen how in random sequences MOT-DETR-2D significantly outperforms FairMOT. This is expected, as FairMOT is designed to work in high frame sequences with overlap between frames, and this does not always happen in the Real-Random sequences.

\begin{table}[htbp]
\centering
\caption{Tracking performance of sorted sequences with small step size between viewpoints. The results of the t-test (p = 0.05) are summarized by different letters sorted in alphabetic order, where group A outperforms B, and B outperforms C.}
\resizebox{\linewidth}{!}{
    \begin{tabular}{cccccccc}
    \hline
    \textbf{Sequence type} & \textbf{Model} & \textbf{HOTA}$\uparrow$ & \textbf{LocA}$\uparrow$ & \textbf{DetA}$\uparrow$ & \textbf{AssA}$\uparrow$ & \textbf{MOTA}$\uparrow$ & \textbf{IDSW}$\downarrow$ \\ \hline
    \multirow{3}{*}{Real-Sort} & MOT-DETR-3D & \textbf{60.4}$^{a}$ & 79.17$^{b}$ & \textbf{56.3}$^{a}$ & \textbf{65.17}$^{a}$ & \textbf{70.38}$^{a}$ & \textbf{23.8}$^{a}$ \\
    & MOT-DETR-2D & 45.22$^{b}$ & 76.13$^{c}$ & 49.83$^{b}$ & 41.64$^{b}$ & 52.38$^{b}$ & 56.2$^{b}$ \\
    & FairMOT \citep{zhang_fairmot_2021} & 46.06$^{b}$ & \textbf{88.87}$^{a}$ & 48.81$^{b}$ & 43.67$^{b}$ & 51.49$^{b}$ & 75.6$^{c}$ \\ 
    \hline
    \multirow{3}{*}{Real-Random} & MOT-DETR-3D & \textbf{59.63}$^{a}$ & 79.14$^{b}$ & \textbf{56.32}$^{a}$ & \textbf{63.44}$^{a}$ & \textbf{66.85}$^{a}$ & 57.4$^{b}$ \\
    & MOT-DETR-2D & 42.83$^{b}$ & 76.09$^{c}$ & 49.64$^{b}$ & 37.58$^{b}$ & 39.27$^{b}$ & 181.0$^{c}$ \\
    & FairMOT \citep{zhang_fairmot_2021} & 31.38$^{c}$ & \textbf{87.89}$^{a}$ & 29.33$^{c}$ & 33.81$^{c}$ & 36.54$^{c}$ & \textbf{48.2}$^{a}$ \\ 
    \hline
    \multirow{2}{*}{Synthetic-Random} & MOT-DETR-3D & \textbf{59.88}$^{a}$ & 81.26$^{a}$ & 62.58$^{b}$ & \textbf{57.46}$^{a}$ & \textbf{65.96}$^{a}$ & \textbf{301.2}$^{a}$ \\
    & MOT-DETR-2D & 56.74$^{b}$ & 81.39$^{a}$ & \textbf{64.2}$^{a}$ & 50.35$^{b}$ & 60.7$^{b}$ & 523.6$^{b}$ \\ \hline
    \end{tabular}
}
\label{tab:combined}
\end{table}

In sequences with more DoF, like the Synthetic-Random experiment, MOT-DETR-3D's performance resembles the Real-Random one. Against Real-Random, the HOTA score only improves 0.25, and the MOTA score decreases 0.89. This suggest that the 3D variant of MOT-DETR optimally leverages 3D data, maintaining consistent performance across varying sequence conditions. In this situation with more DoF the performance decrease might be larger if real data were to be used, as the difference between consecutive viewpoints is larger than in Real-Random. However, the use of synthetic data might counter the performance drop from using more DoF as synthetic data might be easier to process.

\subsection{Ablation studies}
The previous experiment shows the benefit of using the 3D part on MOT-DETR, as it outperforms the 2D version. Additionally, we performed further ablation studies to investigate the effect of using different CNN encoder networks for both the images and the point clouds and the effect of using our synthetic data generated through an L-system-based formalism. We show the results of our experiments in Tables \ref{tab:ablation_cnn} and \ref{tab:ablation_dataset}.

Table \ref{tab:ablation_cnn} presents the tracking performance of our model using the Real-random sequence type with three different CNNs: ResNet18, ResNet34, and ResNet50. The metrics evaluated are HOTA, LocA, DetA, and AssA. The results indicate that the choice of CNN significantly impacts the model's performance. Specifically, ResNet34 achieves the highest overall HOTA, LocA, DetA, and AssA scores when trained on both synthetic and real datasets, outperforming both ResNet50 and ResNet18. ResNet50 is a deeper CNN that outperforms ResNet34 in image classification tasks, and similar behavior could be expected when used in MOT-DETR. However, this is not the case as MOT-DETR with ResNet34 yields the best results. This can be due the fact that MOT-DETR, like DETR, only uses the output of the last convolutional block of the CNNs. The resulting feature map might contain lower spatial resolution features when ResNet50 is used compared to ResNet34. Additinally, it could be that a larger model like ResNet50 requires more training data to achieve optimal performance.

\begin{table}[htbp]
\centering
\caption{Ablation study showing the tracking performance using the Real-random sequence type. Three different CNNs are evaluated: ResNet18, ResNet34 and ResNet50.}
\resizebox{\linewidth}{!}{
    \begin{tabular}{ccccccc}
    \hline
    \textbf{Model} & \textbf{CNNs} & \textbf{Dataset} & \textbf{HOTA} & \textbf{LocA} & \textbf{DetA} & \textbf{AssA} \\
    \hline
    \multirow{3}{*}{MOT-DETR-3D} & Resnet18 & Synthetic+Real & 43.67$^{b}$ & 72.19$^{c}$ & 41.86$^{c}$ & 46.10$^{b}$ \\
    & ResNet34 & Synthetic+Real & \textbf{59.63}$^{a}$ & \textbf{79.14}$^{a}$ & \textbf{56.32}$^{a}$ & \textbf{63.44}$^{a}$ \\
    & ResNet50 & Synthetic+Real & \textbf{57.53}$^{a}$ & 76.29$^{b}$ & 52.59$^{b}$ & \textbf{63.25}$^{a}$ \\
    \hline
    \end{tabular}
}
\label{tab:ablation_cnn}
\end{table}

In Table \ref{tab:ablation_dataset}, we compare the tracking performance of MOT-DETR-3D and MOT-DETR-2D using both a combination of synthetic and real datasets and only the real dataset. ResNet34 serves as the CNN encoder for both models. The results indicate that training with a combination of synthetic and real datasets significantly enhances performance across all metrics compared to training with only real datasets. Moreover, training solely on the real dataset yields poor performance. Despite using pre-trained CNNs, the transformer in MOT-DETR is trained from scratch, and our real training dataset, comprising only 3,750 images, is insufficient for effective training. This highlights the advantages of our L-system-based formalism approach for synthetic data generation, which allows the generation of synthetic training images and point clouds by building 3D models of tomato plants. Furthermore, these results suggest that MOT-DETR requires large datasets to fully reach optimal performance. In this work, we present a method for generating synthetic data that successfully improve the performance of MOT-DETR. However this might not be feasible in all robotic applications.

\begin{table}[htbp]
\centering
\caption{Ablation study showing the tracking performance using the Real-random sequence type. Both MOT-DETR-3D and MOT-DETR-2D were trained using both only our real dataset and a combination of synthetic and real datasets.}
\resizebox{\linewidth}{!}{
    \begin{tabular}{ccccccccc}
    \hline
    \textbf{Model} & \textbf{CNNs} & \textbf{Dataset} & \textbf{HOTA} & \textbf{LocA} & \textbf{DetA} & \textbf{AssA} \\
    \hline
    \multirow{2}{*}{MOT-DETR-3D} & ResNet34 & Synthetic+Real & \textbf{59.63}$^{a}$ & \textbf{79.14}$^{a}$ & \textbf{56.32}$^{a}$ & \textbf{63.44}$^{a}$ \\
    & ResNet34 & Real           & 13.05$^{c}$ & 63.22$^{d}$ & 17.35$^{c}$ & 10.04$^{c}$ \\
    \multirow{2}{*}{MOT-DETR-2D} & ResNet34 & Synthetic+Real & 42.83$^{b}$ & 76.09$^{b}$ & 49.64$^{b}$ & 37.58$^{b}$ \\
    & ResNet34 & Real           & 12.90$^{c}$ & 63.79$^{c}$ & 18.23$^{c}$ & 9.40$^{c}$ \\
    \hline
    \end{tabular}
}
\label{tab:ablation_dataset}
\end{table}

\subsection{Effect of camera pose noise}
Our proposed algorithm takes as input a point cloud that is transformed to the robot world frame using the camera pose at every viewpoint. The data used in our experiments have low camera pose noise since the synthetic data provides perfect camera poses and the robot arm provides highly accurate camera poses. This is not always the case in robotic applications. Robots using odometry or SLAM algorithms are common. These approaches have a higher pose noise. Therefore, to study the effect of camera noise in our algorithm, we added 6DoF artificial Gaussian noise to the camera pose at each viewpoint. 

Table \ref{tab:noise} shows the performance of our algorithm with three different noise levels. The results are displayed as a delta over the same sequences shown in Table \ref{tab:combined} for Real-Random and Synthetic-Random. It can be seen how the performance of MOT-DETR does not present a statistically significant decrease in the Real-Random sequences with the applied noise levels. The only statistically significant decrease can be found when $T_{noise}$ is set to 0.05 in the Synthetic-Random sequences. This suggests that MOT-DETR is resilient to some amount of camera pose inaccuracies. This resilience can be explained by the fact that similar random noise was applied as data augmentation at training time. Furthermore, the real dataset inherently contained camera pose noise due to the nature of the real world system, while the camera pose error on the synthetic dataset was zero. This can explain why with larger noise levels, the performance decreases significantly in the synthetic data. However, MOT-DETR was not evaluated on camera poses generated from SLAM or odometry-based approaches, that can generate larger noise levels that can accumulate over time.

\begin{table}[htbp]
\centering
\caption{Effect of camera pose noise in the performance of MOT-DETR-3D on the sequences of the Real-Random experiment. Values with a * present a significant difference.}
\resizebox{\linewidth}{!}{
    \begin{tabular}{cccccccc} \hline 
    \textbf{Sequence type} & \textbf{T$_{noise}$} & \textbf{$\Delta$ HOTA} & \textbf{$\Delta$ LocA} & \textbf{$\Delta$ DetA} & \textbf{$\Delta$ AssA} & \textbf{$\Delta$ MOTA} & \textbf{$\Delta$ IDSW} \\ \hline 
    \multirow{3}{*}{Real-Random} & 0.001 & 0.17 & 0.01 & -0.02 & 0.38 & -0.12 & 1.40 \\ 
    & 0.01 & 0.13 & 0.00 & -0.07 & 0.35 & -0.39 & 5.80 \\ 
    & 0.05 & -1.88 & -0.07 & 0.12 & -3.89 & -1.95 & 17.20 \\ \hline 
    \multirow{3}{*}{Synthetic-Random} & 0.001 & -0.24 & -0.01 & -0.02 & -0.46 & -0.48 & 14.00 \\ 
    & 0.01 & -0.11 & -0.04 & -0.03 & -0.19 & -0.31 & 8.00 \\ 
    & 0.05 & -4.84$^{*}$ & -0.11 & -0.20 & -8.80$^{*}$ & -12.26$^{*}$ & 386.60$^{*}$ \\ \hline
    \end{tabular}
}
\label{tab:noise}
\end{table}

\section{Conclusion}
In this work, we have introduced MOT-DETR, an algorithm to perform 3D multi-object tracking in robotic multi-view perception applications. 
We showed how MOT-DETR outperformed state-of-the-art MOT algorithms, and can be used by robots to build a representation in real-world challenging environments like a tomato greenhouse. Furthermore, we showed that MOT-DETR can successfully use 3D information to improve 3D MOT tracking in complex sequences with large distance between frames and long-term occlusions. Additionally, we showed that our algorithm is resilient to noise in the camera pose.

\section*{CRediT author statement}
\textbf{David Rapado-Rincon}: Conceptualization, Methodology, Software, Investigation, Data Curation, Writing - Original draft; \textbf{Henk Nap}: Data collection, SOTA comparisons, Writing - Review \& Editing; \textbf{Katarina Smolenova}: Synthetic data collection, Writing - Review \& Editing; \textbf{Eldert J. van Henten}: Conceptualization, Writing - Review \& Editing, Supervision, Funding acquisition; \textbf{Gert Kootstra}: Conceptualization, Writing - Review \& Editing, Supervision, Funding acquisition.

\section*{Funding}
This research is part of the project Cognitive Robotics for Flexible Agro-food Technology (FlexCRAFT), funded by the Netherlands Organization for Scientific Research (NWO) grant P17-01.

\section*{Declaration of competing interest}
We declare that there are no personal and/or financial relationships that have inappropriately affected or influenced the work presented in this paper.

\bibliographystyle{elsarticle-num}  
\bibliography{references}

\begin{thebibliography}{10}
\expandafter\ifx\csname url\endcsname\relax
  \def\url#1{\texttt{#1}}\fi
\expandafter\ifx\csname urlprefix\endcsname\relax\def\urlprefix{URL }\fi
\expandafter\ifx\csname href\endcsname\relax
  \def\href#1#2{#2} \def\path#1{#1}\fi

\bibitem{kootstra_selective_2021}
G.~Kootstra, X.~Wang, P.~M. Blok, J.~Hemming, E.~van Henten,
  \href{https://doi.org/10.1007/s43154-020-00034-1}{Selective {Harvesting}
  {Robotics}: {Current} {Research}, {Trends}, and {Future} {Directions}},
  Current Robotics Reports 2~(1) (2021) 95--104.
\newblock \href {https://doi.org/10.1007/s43154-020-00034-1}
  {\path{doi:10.1007/s43154-020-00034-1}}.
\newline\urlprefix\url{https://doi.org/10.1007/s43154-020-00034-1}

\bibitem{crowley_dynamic_1985}
J.~Crowley, \href{http://ieeexplore.ieee.org/document/1087380/}{Dynamic world
  modeling for an intelligent mobile robot using a rotating ultra-sonic ranging
  device}, in: Proceedings. 1985 {IEEE} {International} {Conference} on
  {Robotics} and {Automation}, Vol.~2, Institute of Electrical and Electronics
  Engineers, St. Louis, MO, USA, 1985, pp. 128--135.
\newblock \href {https://doi.org/10.1109/ROBOT.1985.1087380}
  {\path{doi:10.1109/ROBOT.1985.1087380}}.
\newline\urlprefix\url{http://ieeexplore.ieee.org/document/1087380/}

\bibitem{elfring_semantic_2013}
J.~Elfring, S.~van~den Dries, M.~van~de Molengraft, M.~Steinbuch,
  \href{https://linkinghub.elsevier.com/retrieve/pii/S0921889012002163}{Semantic
  world modeling using probabilistic multiple hypothesis anchoring}, Robotics
  and Autonomous Systems 61~(2) (2013) 95--105.
\newblock \href {https://doi.org/10.1016/j.robot.2012.11.005}
  {\path{doi:10.1016/j.robot.2012.11.005}}.
\newline\urlprefix\url{https://linkinghub.elsevier.com/retrieve/pii/S0921889012002163}

\bibitem{arad_development_2020}
B.~Arad, J.~Balendonck, R.~Barth, O.~Ben‐Shahar, Y.~Edan, T.~Hellström,
  J.~Hemming, P.~Kurtser, O.~Ringdahl, T.~Tielen, B.~v. Tuijl,
  \href{https://www.onlinelibrary.wiley.com/doi/abs/10.1002/rob.21937}{Development
  of a sweet pepper harvesting robot}, Journal of Field Robotics n/a~(n/a),
  \_eprint: https://onlinelibrary.wiley.com/doi/pdf/10.1002/rob.21937 (2020).
\newblock \href {https://doi.org/10.1002/rob.21937}
  {\path{doi:10.1002/rob.21937}}.
\newline\urlprefix\url{https://www.onlinelibrary.wiley.com/doi/abs/10.1002/rob.21937}

\bibitem{burusa_efficient_2023}
A.~K. Burusa, J.~Scholten, D.~R. Rincon, X.~Wang, E.~J. van Henten,
  G.~Kootstra, \href{http://arxiv.org/abs/2306.09801}{Efficient {Search} and
  {Detection} of {Relevant} {Plant} {Parts} using {Semantics}-{Aware} {Active}
  {Vision}}, arXiv:2306.09801 [cs] (Jun. 2023).
\newblock \href {https://doi.org/10.48550/arXiv.2306.09801}
  {\path{doi:10.48550/arXiv.2306.09801}}.
\newline\urlprefix\url{http://arxiv.org/abs/2306.09801}

\bibitem{wong_data_2015}
L.~L. Wong, L.~P. Kaelbling, T.~Lozano-Pérez,
  \href{https://doi.org/10.1177/0278364914559754}{Data association for semantic
  world modeling from partial views}, The International Journal of Robotics
  Research 34~(7) (2015) 1064--1082, publisher: SAGE Publications Ltd STM.
\newblock \href {https://doi.org/10.1177/0278364914559754}
  {\path{doi:10.1177/0278364914559754}}.
\newline\urlprefix\url{https://doi.org/10.1177/0278364914559754}

\bibitem{persson_semantic_2020}
A.~Persson, P.~Z.~D. Martires, A.~Loutfi, L.~De~Raedt,
  \href{http://arxiv.org/abs/1902.09937}{Semantic {Relational} {Object}
  {Tracking}}, IEEE Transactions on Cognitive and Developmental Systems 12~(1)
  (2020) 84--97, arXiv: 1902.09937.
\newblock \href {https://doi.org/10.1109/TCDS.2019.2915763}
  {\path{doi:10.1109/TCDS.2019.2915763}}.
\newline\urlprefix\url{http://arxiv.org/abs/1902.09937}

\bibitem{rapado-rincon_development_2023}
D.~Rapado-Rincón, E.~J. van Henten, G.~Kootstra,
  \href{https://www.sciencedirect.com/science/article/pii/S1537511023001162}{Development
  and evaluation of automated localisation and reconstruction of all fruits on
  tomato plants in a greenhouse based on multi-view perception and {3D}
  multi-object tracking}, Biosystems Engineering 231 (2023) 78--91.
\newblock \href {https://doi.org/10.1016/j.biosystemseng.2023.06.003}
  {\path{doi:10.1016/j.biosystemseng.2023.06.003}}.
\newline\urlprefix\url{https://www.sciencedirect.com/science/article/pii/S1537511023001162}

\bibitem{bewley_simple_2016}
A.~Bewley, Z.~Ge, L.~Ott, F.~Ramos, B.~Upcroft,
  \href{http://arxiv.org/abs/1602.00763}{Simple {Online} and {Realtime}
  {Tracking}}, 2016 IEEE International Conference on Image Processing (ICIP)
  (2016) 3464--3468ArXiv: 1602.00763.
\newblock \href {https://doi.org/10.1109/ICIP.2016.7533003}
  {\path{doi:10.1109/ICIP.2016.7533003}}.
\newline\urlprefix\url{http://arxiv.org/abs/1602.00763}

\bibitem{wojke_simple_2017}
N.~Wojke, A.~Bewley, D.~Paulus, Simple online and realtime tracking with a deep
  association metric, in: 2017 {IEEE} {International} {Conference} on {Image}
  {Processing} ({ICIP}), 2017, pp. 3645--3649, iSSN: 2381-8549.
\newblock \href {https://doi.org/10.1109/ICIP.2017.8296962}
  {\path{doi:10.1109/ICIP.2017.8296962}}.

\bibitem{zhang_fairmot_2021}
Y.~Zhang, C.~Wang, X.~Wang, W.~Zeng, W.~Liu,
  \href{https://doi.org/10.1007/s11263-021-01513-4}{{FairMOT}: {On} the
  {Fairness} of {Detection} and {Re}-identification in {Multiple} {Object}
  {Tracking}}, International Journal of Computer Vision 129~(11) (2021)
  3069--3087.
\newblock \href {https://doi.org/10.1007/s11263-021-01513-4}
  {\path{doi:10.1007/s11263-021-01513-4}}.
\newline\urlprefix\url{https://doi.org/10.1007/s11263-021-01513-4}

\bibitem{meinhardt_trackformer_2022}
T.~Meinhardt, A.~Kirillov, L.~Leal-Taixe, C.~Feichtenhofer,
  \href{https://ieeexplore.ieee.org/document/9879668/}{{TrackFormer}:
  {Multi}-{Object} {Tracking} with {Transformers}}, in: 2022 {IEEE}/{CVF}
  {Conference} on {Computer} {Vision} and {Pattern} {Recognition} ({CVPR}),
  IEEE, New Orleans, LA, USA, 2022, pp. 8834--8844.
\newblock \href {https://doi.org/10.1109/CVPR52688.2022.00864}
  {\path{doi:10.1109/CVPR52688.2022.00864}}.
\newline\urlprefix\url{https://ieeexplore.ieee.org/document/9879668/}

\bibitem{zeng_motr_2022}
F.~Zeng, B.~Dong, Y.~Zhang, T.~Wang, X.~Zhang, Y.~Wei,
  \href{http://arxiv.org/abs/2105.03247}{{MOTR}: {End}-to-{End}
  {Multiple}-{Object} {Tracking} with {Transformer}}, arXiv:2105.03247 [cs]
  (Jul. 2022).
\newblock \href {https://doi.org/10.48550/arXiv.2105.03247}
  {\path{doi:10.48550/arXiv.2105.03247}}.
\newline\urlprefix\url{http://arxiv.org/abs/2105.03247}

\bibitem{carion_end--end_2020}
N.~Carion, F.~Massa, G.~Synnaeve, N.~Usunier, A.~Kirillov, S.~Zagoruyko,
  \href{http://arxiv.org/abs/2005.12872}{End-to-{End} {Object} {Detection} with
  {Transformers}}, arXiv:2005.12872 [cs] (May 2020).
\newblock \href {https://doi.org/10.48550/arXiv.2005.12872}
  {\path{doi:10.48550/arXiv.2005.12872}}.
\newline\urlprefix\url{http://arxiv.org/abs/2005.12872}

\bibitem{halstead_fruit_2018}
M.~Halstead, C.~McCool, S.~Denman, T.~Perez, C.~Fookes,
  \href{https://ieeexplore.ieee.org/document/8392450/}{Fruit {Quantity} and
  {Ripeness} {Estimation} {Using} a {Robotic} {Vision} {System}}, IEEE Robotics
  and Automation Letters 3~(4) (2018) 2995--3002.
\newblock \href {https://doi.org/10.1109/LRA.2018.2849514}
  {\path{doi:10.1109/LRA.2018.2849514}}.
\newline\urlprefix\url{https://ieeexplore.ieee.org/document/8392450/}

\bibitem{kirk_robust_2021}
R.~Kirk, M.~Mangan, G.~Cielniak, Robust {Counting} of {Soft} {Fruit} {Through}
  {Occlusions} with {Re}-identification, in: M.~Vincze, T.~Patten, H.~I.
  Christensen, L.~Nalpantidis, M.~Liu (Eds.), Computer {Vision} {Systems},
  Lecture {Notes} in {Computer} {Science}, Springer International Publishing,
  Cham, 2021, pp. 211--222.
\newblock \href {https://doi.org/10.1007/978-3-030-87156-7_17}
  {\path{doi:10.1007/978-3-030-87156-7_17}}.

\bibitem{halstead_crop_2021}
M.~Halstead, A.~Ahmadi, C.~Smitt, O.~Schmittmann, C.~McCool,
  \href{https://www.frontiersin.org/article/10.3389/fpls.2021.786702}{Crop
  {Agnostic} {Monitoring} {Driven} by {Deep} {Learning}}, Frontiers in Plant
  Science 12 (2021).
\newblock \href {https://doi.org/10.3389/fpls.2021.786702}
  {\path{doi:10.3389/fpls.2021.786702}}.
\newline\urlprefix\url{https://www.frontiersin.org/article/10.3389/fpls.2021.786702}

\bibitem{villacres_apple_2023}
J.~Villacrés, M.~Viscaino, J.~Delpiano, S.~Vougioukas, F.~Auat~Cheein,
  \href{https://www.sciencedirect.com/science/article/pii/S0168169922008213}{Apple
  orchard production estimation using deep learning strategies: {A} comparison
  of tracking-by-detection algorithms}, Computers and Electronics in
  Agriculture 204 (2023) 107513.
\newblock \href {https://doi.org/10.1016/j.compag.2022.107513}
  {\path{doi:10.1016/j.compag.2022.107513}}.
\newline\urlprefix\url{https://www.sciencedirect.com/science/article/pii/S0168169922008213}

\bibitem{rapado-rincon_minksort_2023}
D.~Rapado-Rincón, E.~J. van Henten, G.~Kootstra,
  \href{https://www.sciencedirect.com/science/article/pii/S1537511023002374}{{MinkSORT}:
  {A} {3D} deep feature extractor using sparse convolutions to improve {3D}
  multi-object tracking in greenhouse tomato plants}, Biosystems Engineering
  236 (2023) 193--200.
\newblock \href {https://doi.org/10.1016/j.biosystemseng.2023.11.003}
  {\path{doi:10.1016/j.biosystemseng.2023.11.003}}.
\newline\urlprefix\url{https://www.sciencedirect.com/science/article/pii/S1537511023002374}

\bibitem{loshchilov_decoupled_2019}
I.~Loshchilov, F.~Hutter, \href{http://arxiv.org/abs/1711.05101}{Decoupled
  {Weight} {Decay} {Regularization}}, arXiv:1711.05101 [cs, math] (Jan. 2019).
\newblock \href {https://doi.org/10.48550/arXiv.1711.05101}
  {\path{doi:10.48550/arXiv.1711.05101}}.
\newline\urlprefix\url{http://arxiv.org/abs/1711.05101}

\bibitem{hemmerling_rule-based_2008}
R.~Hemmerling, O.~Kniemeyer, D.~Lanwert, W.~Kurth, G.~Buck-Sorlin,
  \href{https://www.publish.csiro.au/fp/FP08052}{The rule-based language {XL}
  and the modelling environment {GroIMP} illustrated with simulated tree
  competition}, Functional Plant Biology 35~(10) (2008) 739--750, publisher:
  CSIRO PUBLISHING.
\newblock \href {https://doi.org/10.1071/FP08052} {\path{doi:10.1071/FP08052}}.
\newline\urlprefix\url{https://www.publish.csiro.au/fp/FP08052}

\bibitem{zhou_open3d_2018}
Q.-Y. Zhou, J.~Park, V.~Koltun,
  \href{http://arxiv.org/abs/1801.09847}{{Open3D}: {A} {Modern} {Library} for
  {3D} {Data} {Processing}}, arXiv:1801.09847 [cs] (Jan. 2018).
\newblock \href {https://doi.org/10.48550/arXiv.1801.09847}
  {\path{doi:10.48550/arXiv.1801.09847}}.
\newline\urlprefix\url{http://arxiv.org/abs/1801.09847}

\bibitem{afonso_tomato_2020}
M.~Afonso, H.~Fonteijn, F.~S. Fiorentin, D.~Lensink, M.~Mooij, N.~Faber,
  G.~Polder, R.~Wehrens,
  \href{https://www.frontiersin.org/article/10.3389/fpls.2020.571299}{Tomato
  {Fruit} {Detection} and {Counting} in {Greenhouses} {Using} {Deep}
  {Learning}}, Frontiers in Plant Science 11 (2020).
\newblock \href {https://doi.org/10.3389/fpls.2020.571299}
  {\path{doi:10.3389/fpls.2020.571299}}.
\newline\urlprefix\url{https://www.frontiersin.org/article/10.3389/fpls.2020.571299}

\end{thebibliography}

\end{document}